%% file: acl_latex.tex
\documentclass[11pt]{article}

\usepackage[final]{acl}

\usepackage{times}
\usepackage{latexsym}

\usepackage[T1]{fontenc}

\usepackage[utf8]{inputenc}

\usepackage{microtype}

\usepackage{inconsolata}

\usepackage{graphicx}
\usepackage{amssymb}
\usepackage{booktabs}
\usepackage{tabularx}
\usepackage{multirow}
\usepackage[table]{xcolor}
\usepackage{soul}
\usepackage{amsmath}
\usepackage{booktabs}
\usepackage{makecell}
\usepackage{cuted}    
\usepackage{caption}  

\title{\raisebox{-0.5ex}{\protect\includegraphics[clip, height=2.5\fontcharht\font`\B]{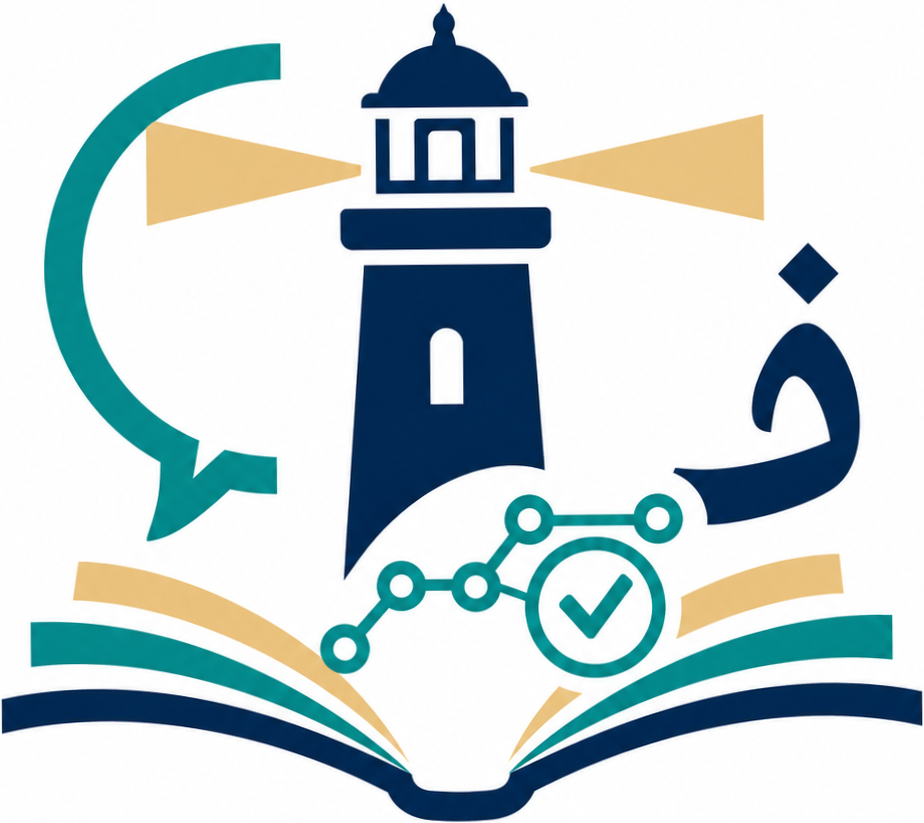}} AlexandriaX 2026:\\The First Shared Task on Dialectal Arabic Machine Translation
\vspace{0.2em}}

\author{
    \textbf{Abdellah {El Mekki}}$^{\lambda}$ \qquad
    \textbf{AbdelRahim {A. Elmadany}}$^{\lambda}$ \qquad
    \textbf{Samar {M. Magdy}}$^{\lambda}$ \\
    \textbf{Saad Ezzini}$^{\alpha}$ \qquad
    \textbf{Mo El-Haj}$^{\beta,\eta}$ \qquad
    \textbf{Mustafa Jarrar}$^{\gamma}$ \\
    \textbf{Zaid Alyafeai}$^{\delta}$ \qquad
    \textbf{Bernard Ghanem}$^{\delta}$ \qquad
    \textbf{Muhammad Abdul-Mageed}$^{\lambda,\kappa}$ \\[2ex]
    \normalfont \normalsize
    $^{\lambda}$The University of British Columbia \quad
    $^{\kappa}$Canada Research Chair in NLP and ML \\
    $^{\alpha}$King Fahd University of Petroleum and Minerals \quad
    $^{\beta}$Lancaster University \quad
    $^{\eta}$VinUniversity \\
    $^{\gamma}$Hamad Bin Khalifa University \quad
    $^{\delta}$King Abdullah University of Science and Technology
    \\ \texttt{\{abdellah.elmekki, muhammad.mageed\}@ubc.ca}
}

\begin{document}
\maketitle

\begin{strip}
  \centering
  \includegraphics[trim={0cm 0cm 0cm 0cm}, clip, width=\textwidth]{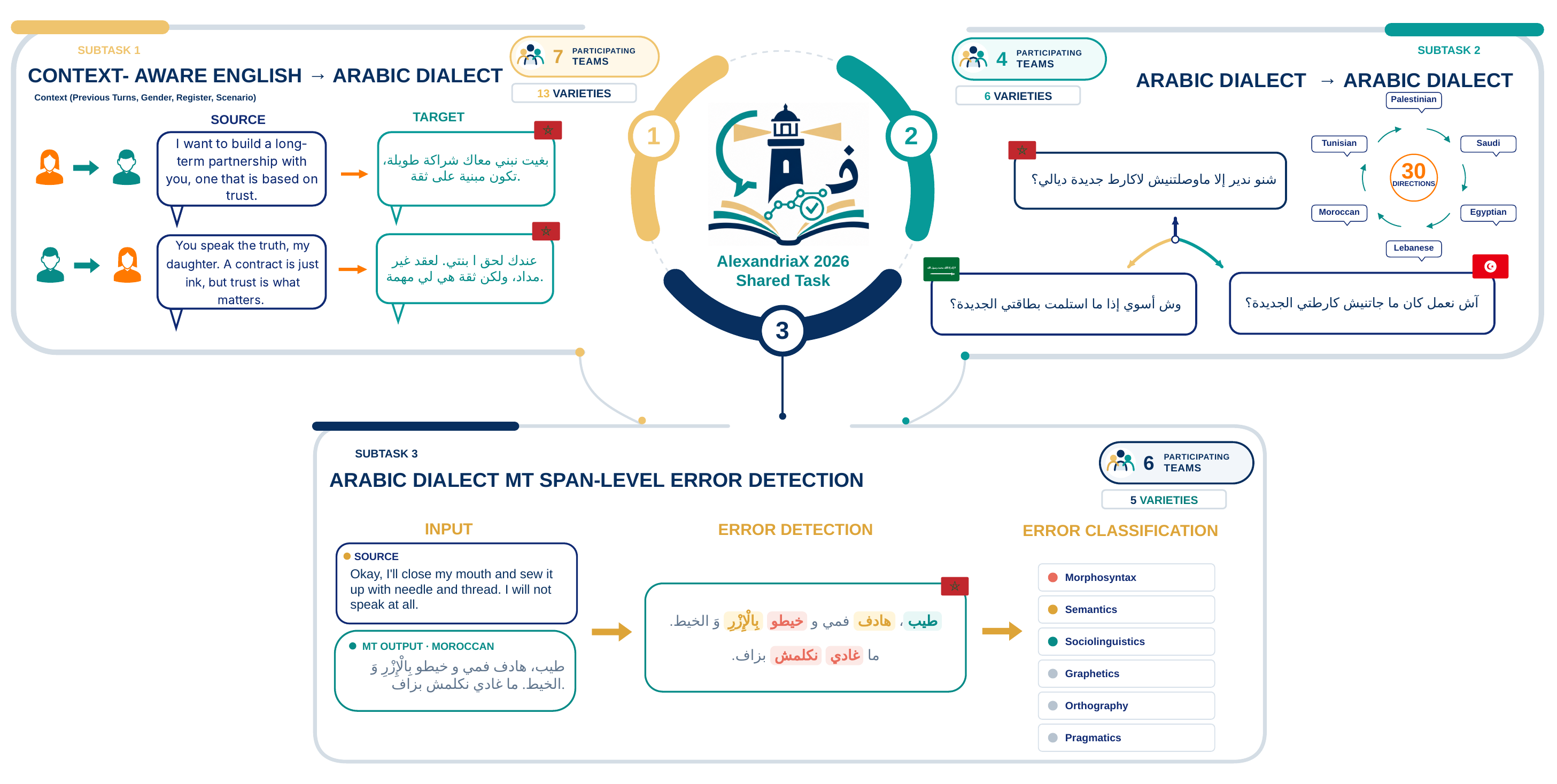}
  \captionof{figure}{Overview of the three AlexandriaX 2026 subtasks: context-aware English-to-Dialectal Arabic translation, cross-dialect Arabic translation, and span-level detection and classification of translation errors.}
  \label{fig:main_figure}
\end{strip}

\begin{abstract}
Dialectal Arabic machine translation (MT) remains challenging despite recent progress in Arabic language technologies, particularly because effective translation requires modeling not only semantic content but also dialectal variation, conversational context, speaker and addressee characteristics, and sociolinguistic appropriateness. Moreover, conventional MT metrics provide limited insight into the linguistic errors produced by dialectal systems. We present the AlexandriaX~2026 Shared Task on Dialectal Arabic MT, which addresses these challenges through three complementary subtasks: (1) context-aware English-to-Dialectal Arabic dialogue translation across 13 Arabic varieties, (2) cross-dialect Arabic translation in the financial domain covering six Arabic dialects, and (3) span-level MT error detection and classification using linguistically motivated error categories across five Arabic varieties. The shared task attracted 38 registrations for Subtask~1, 33 for Subtask~2, and 35 for Subtask~3. Twelve unique teams submitted their system description papers, all of which we accepted for publication. The best system on Subtask~1 achieved 30.42 spBLEU in the constrained track and 33.49 spBLEU in the unconstrained track. On Subtask~2, the top system obtained 28.40 spBLEU. 
On Subtask~3, the best system achieved an overall score of 49.82, outperforming the 24.29 baseline. Taken together, the results of the leading systems across the three subtasks highlight the benefits of explicit dialect modeling, context-aware generation, retrieval and reranking, and specialized approaches to interpretable MT error analysis. All the resources of AlexandriaX~2026 shared task are publicly available, including data, baselines, and evaluation code on our project page: \url{https://alexandriax.dlnlp.ai}. 
\end{abstract}

\input{sections/introduction}
\input{sections/task_description}
\input{sections/rules_eval}
\input{sections/teams_results}
\input{sections/discussion}

\input{sections/conclusion}

\input{sections/limitations}

\input{sections/acknowledgments}

\bibliography{custom}

\appendix
\section*{Appendices}

\input{sections/appendix}

\end{document}

%% file: sections/introduction.tex
\section{Introduction}

Machine translation (MT) for Arabic has improved substantially in recent years, driven by advances in neural models and large language models (LLMs) \cite{kadaoui-etal-2023-tarjamat}. However, this progress remains uneven across Arabic varieties. Most resources and modeling efforts have historically focused on Modern Standard Arabic (MSA) \cite{ bergman-diab-2022-towards, elmadany2022turjuman, elmadany2023octopus}, while current systems remain considerably less reliable on Dialectal Arabic (DA) and often struggle to generalize across dialects and sub-dialects \cite{khalifa-etal-2016-large,el-mekki-etal-2026-alexandria}. This gap is especially important because dialects are the primary medium of everyday communication across the Arab world and exhibit substantial variation across regions and communities \cite{zaidan-callison-burch-2014-arabic,bouamor-etal-2018-madar}.

Dialectal Arabic introduces challenges that go beyond standard sentence-level translation. Linguistic choices can depend on the target dialect, speaker and addressee gender, register, domain, persona, and preceding dialogue context. A translation may therefore preserve the source meaning while still sounding unnatural or inappropriate because of incorrect gender agreement, unsuitable formality, dialect mismatches, or the use of MSA constructions where dialectal forms are expected. High-quality dialectal MT consequently requires models that can account for conversational and sociolinguistic context and generalize across dialects with different levels of resource availability~\citep{el-mekki-etal-2026-alexandria}.

Improving translation quality alone is not sufficient; it is also necessary to understand how translation systems fail. Common automatic MT metrics provide useful aggregate scores but offer limited insight into the linguistic nature of errors~\cite{popovic-ney-2011-towards}. This limitation is particularly important for dialectal Arabic, where translation errors may involve semantics, morphology, orthography, pragmatics, or sociolinguistic appropriateness~\citep{magdy-etal-2026-lqm}. Manual error analysis can capture these distinctions, but it is costly and difficult to scale.

There is therefore a need for automatic and interpretable translation-quality analysis that can identify erroneous spans and characterize the types of errors being made. The Linguistically Motivated Multidimensional Quality Metrics (LQM) framework provides a basis for this goal by organizing translation errors into sociolinguistic, pragmatic, semantic, morphosyntactic, orthographic, and graphetic categories~\citep{magdy-etal-2026-lqm}. 

Motivated by these challenges, the AlexandriaX~2026 Shared Task on Context-Aware Dialectal Arabic MT and MT Evaluation focuses on context-aware English-to-Dialectal Arabic translation, Arabic cross-dialect translation, and automatic span-level translation error detection and classification. Together, these tasks aim to advance MT systems that are dialect-aware, context-sensitive, robust across Arabic varieties, and accompanied by interpretable mechanisms for automatically characterizing their errors.

Across the three subtasks, results highlight the value of dialect-specific adaptation, context-aware generation, retrieval and candidate selection, and hybrid approaches to error analysis.

The remainder of this paper is organized as follows: Section~\ref{sec:shared-task-overview} describes the AlexandriaX 2026 Shared Task, while Section~\ref{sec:rules-eval} outlines the participation rules and evaluation framework. Section~\ref{sec:teams-results} presents the participating teams alongside their results, and Section~\ref{sec:discussion} discusses the findings. Finally, Section~\ref{sec:conclusion} concludes the paper. In addition, Appendix~\ref{sec:related-work} reviews related work in the literature.

%% file: sections/task_description.tex
\section{Task Description: AlexandriaX 2026} \label{sec:shared-task-overview}

The objective of the AlexandriaX 2026 Shared Task is to advance the development and evaluation of Machine Translation (MT) systems for Dialectal Arabic through three complementary subtasks: context-aware English-to-Dialectal Arabic dialogue translation, cross-dialect Arabic MT and dialectal Arabic MT error detection and classification. Figure~\ref{fig:main_figure} provides an overall description of the three subtasks along with a data point example from each subtask. The first subtask requires systems to translate English dialogue turns into one of 13 Arabic varieties while leveraging conversational context and metadata such as domain, speaker and addressee gender, and persona information. 
The second subtask focuses on translating short financial-domain queries across six Arabic varieties, including both seen and unseen dialect pairs, to evaluate cross-dialect generalization. The third subtask focuses on automatic MT error analysis for five Arabic dialects, where systems identify erroneous spans in translated text and classify them into six LQM-inspired linguistic categories.


\subsection{Subtask~1: Context-Aware Dialectal Arabic Dialogue Machine Translation}
The goal of this subtask is to encourage the development of MT systems capable of generating natural and contextually appropriate Dialectal Arabic from English dialogue. Unlike conventional sentence-level MT, systems are expected to make use of the broader conversational context and available metadata, including the target dialect, domain, speaker and addressee gender, and persona information (role, profession). The subtask covers 13 Arabic varieties representing Egypt (EG), Jordan (JO), Lebanon (LB), Libya (LY), Mauritania (MR), Morocco (MA), Oman (OM), Palestine (PS), Saudi Arabia (SA), Sudan (SD), Syria (SY), Tunisia (TN), and Yemen (YE). It further spans 11 domains, including agriculture, commerce, education, healthcare, legal and financial services, logistics, workplace communication, and tourism.

Produced translations should preserve the meaning of the English source while sounding natural in the target dialect and appropriate with respect to lexical, morphological, pragmatic, and sociolinguistic choices. The subtask includes two tracks: a \textit{constrained track}, in which participants may only use the Alexandria data~\cite{el-mekki-etal-2026-alexandria} and systems with at most 5B parameters, and an \textit{unconstrained track}, which allows the use of external data and models without a parameter limit.

\paragraph{Data Collection and Annotation}

The data for this subtask are derived from the \textbf{Alexandria} dataset \citep{el-mekki-etal-2026-alexandria}, a large-scale English-Dialectal Arabic parallel corpus consisting of multi-turn conversational scenarios. The dataset was constructed through a combination of controlled LLM-based source generation and community-driven human translation and revision.

More specifically, the Alexandria data was created through three stages: English conversation generation, human dialectal Arabic translation, and revision and quality control. English dialogues were generated with Gemini-2.5 Pro \cite{comanici2025gemini25pushingfrontier} using country-, domain-, persona-, and gender-specific prompts. Native or primary dialect speakers then translated the dialogues into natural local Arabic, preserving meaning, register, gender, and appropriate code-switching. Finally, a second contributor from the same country reviewed each translation for semantic accuracy, dialectal authenticity, consistency, and linguistic quality.

The final Alexandria resource comprises approximately 107K parallel turns across 13 Arab countries and 11 domains. For the shared task, training and development data are provided to participants for system development and local evaluation. The final evaluation was performed on a private test set covering approximately 1,000 turns per dialect, and the participants had access only to its English source. 
 Importantly, Libyan and Sudanese Arabic are absent from the training and development data but included in the test set, allowing the subtask to evaluate generalization to dialectal varieties unseen during development.

\subsection{Subtask~2: Cross-Dialect Arabic Machine Translation}
The primary objective of Subtask~2 is to advance MT across Arabic dialects. While dialect-English and dialect-MSA MT have received growing attention, direct cross-dialect translation remains an under-resourced and challenging problem. Arabic dialects diverge substantially across regions in phonology, morphology, lexica, and syntax. At the same time, speakers of different dialects frequently interact in business, commerce, and digital services, creating a practical need for direct cross-dialect translation systems that do not rely on English or MSA as an intermediate pivot.

In this subtask, systems translate short queries from a source Arabic variety into a designated target variety. Translations must preserve semantic intent, domain terminology, and numerical entities while using natural target-dialect morphosyntax and vocabulary, without reverting to MSA or simply copying the source.

\paragraph{Data Collection and Setup}
Building upon previous efforts in Arabic financial NLP benchmarks, such as AraFinNLP 2024 \citep{malaysha-etal-2024-arafinnlp}, the benchmark for Subtask~2 is constructed from an expanded multi-dialect version of the \textbf{ArBanking77} dataset \citep{jarrar-etal-2023-arbanking77}, a localized Arabic extension of the Banking77 intent-classification benchmark. The dataset comprises short customer-service queries in the financial and banking domain across MSA and several major regional dialects, including Egyptian, Lebanese, Moroccan, Palestinian, Saudi, and Tunisian Arabic. All queries were localized and translated by native speakers of the corresponding varieties.

The official evaluation was conducted on a blind test set comprising 3,574 instances across 30 directed language-pair combinations (derived from 3,518 distinct source dialectal Arabic queries). To test systems' intrinsic cross-dialect generalization and adaptation abilities, no dedicated parallel training or development splits covering all 30 directed pairs were officially provided. Participants were tasked with developing strategies that generalize across dialects using zero-shot prompting, cross-dialect mining, data augmentation, or multilingual transfer learning.

\subsection{Subtask~3: Machine Translation Error Detection and Classification}

Subtask~3 focuses on fine-grained error analysis of dialectal Arabic MT. Given an English source sentence and its machine-generated translation into a target Arabic dialect, participating systems are required to identify erroneous span(s) in the translation and assign an error category to each detected span. The task follows the Linguistically Motivated Multidimensional Quality Metrics (LQM) framework of~\citet{magdy-etal-2026-lqm}, covering six error categories: \textit{sociolinguistics}, \textit{pragmatics}, \textit{semantics}, \textit{morphosyntax}, \textit{orthography/writing conventions}, and \textit{graphetics}.

\paragraph{Data Split}
The data is derived from the LQM resource~\cite{magdy-etal-2026-lqm} and covers five English-to-dialectal-Arabic translation directions: Egyptian (\texttt{ENG\_EGY}), Emirati (\texttt{ENG\_UAE}), Mauritanian (\texttt{ENG\_MAU}), Moroccan (\texttt{ENG\_MOR}), and Palestinian (\texttt{ENG\_PAL}). The dataset is split into 1,125 training, 138 development, and 145 test instances. Each training and development instance contains an English source sentence, its Arabic MT output, and one or more error annotations represented by character-level \texttt{start} and \texttt{end} offsets together with the corresponding LQM category. For the blind test set, the gold error spans and categories were withheld for official evaluation.

%% file: sections/rules_eval.tex
\section{Rules and Evaluation} \label{sec:rules-eval}

\subsection{Participation and Submission Guidelines}

To ensure fairness during the final submission phase, all registered participants were given access to the training and development data to build and evaluate their models across all subtasks. We refer to this stage as the \textit{development phase}. Subsequently, we released the unlabeled blind test data and gave participants approximately ten days to submit their system predictions. All test-set evaluations were conducted on CodaBench. Except for the constrained track in Subtask 1, participants were permitted to use any models (open- or closed-source) and external resources to build their final systems. Once the test phase concluded, submissions were finalized before the final leaderboard was released. For Subtask 1, participants were asked to specify the track to which they were submitting.


\subsection{Evaluation Method}

For evaluating submissions, we report the following metrics:
For Subtasks~1 and 2, we follow~\citet{el-mekki-etal-2026-alexandria} and use reference-based surface form SacreBLEU metrics: spBLEU (using the FLORES-200 SentencePiece tokenizer) \cite{nllb2022} and chrF++ ($\beta=2$, word order=2) \cite{popovic-2015-chrf}. Participants had access to per-dialect breakdowns of these metrics on CodaBench when submitting their systems.

For Subtask~3, we follow the evaluation framework of~\citet{magdy-etal-2026-lqm}, reporting \textit{Exact Match F\textsubscript{1}}, \textit{Overlap F\textsubscript{1}}, and \textit{Error Class Macro-F\textsubscript{1}}. The first two metrics assess span localization under exact and partial overlap criteria, respectively, while \textit{Error Class Macro-F\textsubscript{1}} evaluates error-category prediction. We additionally introduce an \textit{Overall Score} as the official ranking metric, providing a single measure that jointly captures span localization and error classification. It is computed as:

\[
\textit{Overall Score}
=
\frac{1}{2}
\left(
F_{\mathrm{overlap}} + F_{\mathrm{class}}
\right),
\]

\noindent
where $F_{\mathrm{overlap}}$ denotes \textit{Overlap F\textsubscript{1}} and $F_{\mathrm{class}}$ denotes \textit{Error Class Macro-F\textsubscript{1}}. The two components are weighted equally, favoring systems that perform well on both span detection and error classification.

%% file: sections/teams_results.tex
\section{Shared Task Teams and Results} \label{sec:teams-results}

\subsection{Participating Teams}
 \input{tables/participants}

The AlexandriaX~2026 shared task attracted significant interest, with 38 teams registering for Subtask~1, 33 teams registering for Subtask~2, and 35 teams registering for Subtask~3.
Actual participation rates varied across the subtasks. Seven teams submitted their final predictions for the test phase of Subtask~1, four teams for Subtask~2, and six teams for Subtask~3. Two teams participated in all three subtasks. 
Table~\ref{tab:team-participation} presents the participating teams with references to their submitted system description papers, their affiliations, and the subtasks to which they submitted their systems. In total, 13 unique teams participated and submitted their final systems, of which 12 also submitted system description papers.

\subsection{Baselines}

For each subtask, we provided participants with baseline code and reference benchmark scores. For Subtask~1, we used NileChat-3B~\cite{el-mekki-etal-2025-nilechat}, fine-tuned on the Alexandria training dataset using QLoRA~\cite{3666122.3666563}. 

For Subtask~2, we established two complementary baselines: \textit{Leave-as-is} (Identity Copy) and a zero-shot translation using Gemma-4-31B-IT. More details presented in Appendix \ref{app:baselines}.

Finally, for Subtask~3, we used NileChat-3B fine-tuned using QLoRA to produce structured outputs corresponding to the LQM categories. 

\subsection{Submitted Systems and Results}

\input{tables/results}

\input{tables/subtask1_uncost_breakdown}
\input{tables/subtask1_const_breakdown}
\input{tables/subtask2_breakdown_target_dialect}
\input{tables/subtask3_span_dialect_level}

\input{tables/subtask3_class_dialect_level}

\paragraph{Subtask 1: Context-Aware Dialectal Arabic Dialogue Machine Translation.}
Most systems adapted an Arabic-aware or multilingual pretrained model to the Alexandria data and conditioned translation on the target dialect, dialogue history, and speaker metadata. Parameter-efficient fine-tuning was the dominant approach: Thakaa, LahjaMT, Rosetta, and Alkhder adapted NileChat-3B-Base using LoRA~\cite{hu2022lora} or QLoRA~\cite{3666122.3666563}; FCDS fine-tuned Gemma-3-4B-IT~\cite{gemmateam2025gemma3technicalreport}; CUNI used Gemma-4-31B-IT~\cite{gemmateam2026gemma4technicalreport}; and NAMAA Community evaluated AraT5v2, NLLB-200, NileChat, Qwen, and Gemma
\citep{alamr-etal-2026-thakaa,abdallah-elbeltagy-2026-lahjamt,esmaeil-etal-2026-rosetta,alkhder-etal-2026-context,aly-etal-2026-fcds,jon-bojar-2026-cuni,eldin-etal-2026-namaa}.

Systems differed in how they represented dialogue context. CUNI translated whole conversations jointly, obtaining an approximately 0.7 chrF++ development gain over turn-level translation. FCDS performed best when trained with the two most recent English turns but decoded with the complete source and model-generated target history. In contrast, Alkhder found in a post-submission analysis that self-generated history performed worse than both gold history and no history, illustrating the risk of error propagation. Rosetta addressed this mismatch by corrupting reference histories during training, while LahjaMT found only modest gains from metadata and retrieved demonstrations. Overall, context was beneficial when systems accounted for the difference between clean training histories and noisy model-generated histories at inference time.

Inference-time selection was another major source of performance improvement. Thakaa's unconstrained system used candidate-constrained MBR~\cite{kumar-byrne-2004-minimum}: four systems provided selectable translations, while seven additional systems contributed only pseudo-reference evidence. This yielded 33.49 spBLEU, 1.25 points above the best complete candidate. CUNI applied MBR to a 48-candidate pool, improving from 31.42 to 32.47 spBLEU, while its subsequent genetic optimization added only 0.05 points. FCDS similarly improved from 25.52 to 26.48 by selecting among 50 sampled translations. LahjaMT and NAMAA instead routed each dialect to a selected checkpoint-prompt configuration. These results show that selection can substantially improve translation, but it depends strongly on candidate diversity: several correlated systems may reinforce one another without providing independent evidence.

Dialect-matched supervision was particularly important for Libyan and Sudanese Arabic, which were absent from the original training and development splits. Thakaa routed these dialects to a fold-in LoRA adapter trained with their released public-test examples, increasing constrained-track spBLEU from 29.59 to 30.42. LahjaMT's external-data specialists produced a further 0.24-point gain in the unconstrained track. Broader external-data use was less reliable: Rosetta's MADAR~\cite{bouamor-etal-2018-madar} and PADIC~\cite{meftouh-etal-2015-machine} adaptation improved only Libyan and Moroccan while reducing the overall score, and CUNI's synthetic and WMT-derived data did not improve its strongest standalone checkpoint. Across systems, Mauritanian was frequently among the weakest varieties, while code-switching and non-standardized spelling remained important sources of disagreement with the single references. Overall, the results favor task-specific adaptation, carefully designed context, and targeted dialect supervision over model scale or data volume alone.

The official Subtask~1 results, including the baseline results, are presented in Tables~\ref{tab:alexandriax_subtask1_unconstrained} and \ref{tab:alexandriax_subtask1_constrained}, while Tables~\ref{tab:s1_cons_spbleu_results} and \ref{tab:s1_uncons_spbleu_results} present the country-level breakdown scores for the constrained and unconstrained tracks, respectively.

\paragraph{Subtask 2: Cross-Dialect Arabic Machine Translation.}
A total of four teams submitted predictions for Subtask~2 (Table~\ref{tab:alexandriax_subtask2}), with three teams submitting accompanying system description papers \citep{alamr-etal-2026-thakaa,ibrahim-etal-2026-alexis,eldin-etal-2026-namaa}. Participating systems addressed this Subtask through retrieval-guided generation, adaptation on constructed parallel data, and zero-shot prompting.

Thakaa used a frozen Gemma-4-31B-IT model~\cite{gemmateam2026gemma4technicalreport} and retrieved up to 32 nonparallel examples written in the requested target dialect from the released input pool. The examples supplied target-side vocabulary, morphology, spelling, and style, while a guided prompt asked the model to synthesize a source-faithful translation. Retrieval raised spBLEU from 24.68 to 27.94, and guided synthesis produced the winning score of 28.40. The method was transductive because it used the released test inputs as retrieval evidence. Although reciprocal-neighbor filtering was applied, a retrospective audit found two effective cross-dialect counterparts among the 3,574 prompts.

ALEXIS adapted NLLB-200-distilled-600M~\cite{nllb2022} on 70,139 examples constructed from ArBanking77. The corpus combined human-written MSA--Palestinian pairs, intent-matched but unverified cross-dialect pairs, and synthetic Egyptian and Lebanese data. Increasing adaptation capacity and training duration yielded larger gains than further data expansion, and the official system ranked second with 25.84 spBLEU and 39.02 chrF++. 

NAMAA Community used Gemini-2.5-Flash zero-shot, with prompts specifying the source and target varieties and instructions to preserve meaning while avoiding MSA drift and unnecessary copying. A validation and repair stage handled empty, malformed, truncated, or inappropriate identity outputs. The system ranked third with 24.53 spBLEU and 39.04 chrF++.

The fourth participating team, \textbf{L3IA}, ranked fourth on the official leaderboard with 14.07 spBLEU and 29.75 chrF++, outperforming the Leave-as-is baseline (12.81 spBLEU / 27.31 chrF++).

Together, these systems show that retrieval, mining, synthesis, and prompting can compensate for missing parallel data, but their effectiveness depends on the provenance and reliability of the resulting evidence. Nonparallel text is useful as target-dialect guidance, mined or synthetic pairs provide weak supervision, and only sufficiently reliable data should be treated as reference translations.

Table~\ref{tab:alexandriax_subtask2} presents the official results for the systems submitted to Subtask~2, along with the baselines. A breakdown of the results per target dialect variety is presented in Table~\ref{tab:subtask2_target_dialect}. A further description of the breakdown is provided in Appendix~\ref{app:eval}.

\paragraph{Subtask 3: Machine Translation Error Detection and Classification.}
The six systems submitted to Subtask~3 followed three main approaches: discriminative span tagging, generative structured prediction, and hybrid pipelines combining generative semantic reasoning with encoder-based localization
\citep{alamr-etal-2026-thakaa,eldin-etal-2026-namaa,abusaleh-etal-2026-ttlab,labib-etal-2026-reglat,hossain-2026-axiom,Sarah-2026-ArabicMTDiagnostics}.

TTLab and Axiom formulated the task as MARBERTv2-based~\cite{abdul-mageed-etal-2021-arbert} BIO sequence labeling. This anchored predictions to the input and avoided generating offsets explicitly. TTLab combined focal loss~\cite{focal_loss}, direction-specific thresholds, and light boundary post-processing, ranking third with 40.91 Overall. Axiom used inverse-frequency class weighting and confidence-thresholded decoding, ranking fifth with 39.26. REGLAT separated binary span detection from category classification and ensembled MARBERTv2 and CAMeLBERT-DA through a shared character grid. It ranked fourth with 40.21 Overall, but aggressive gap bridging and dilation produced only 0.41 exact-match F1 despite 42.72 overlap F1, which the authors characterized as a ``dilation trap.''

ArabicMTDiagnostics retained a generative formulation, fine-tuning NileChat-3B-Base with QLoRA to emit JSON spans and categories. Input-side character annotations helped with offset generation, while category definitions and examples addressed minority classes. However, union voting over five sampled predictions over-generated spans, resulting in 34.84 Overall and the lowest overlap F1. These experiments showed that structured generation can support joint reasoning, but remains vulnerable to malformed output, inaccurate offsets, and excessive recall.

The two leading systems combined heterogeneous predictors. Thakaa used GPT-5.5~\cite{openai2026gpt55} to propose semantically motivated error regions and MARBERTv2 to refine their boundaries and corroborate additional spans. Agreement-gated additions, fallback handling, and fragment repair raised Overall from 46.91 for the raw GPT-5.5 output to the winning 49.82, with 54.26 overlap F1 and 45.38 class macro-F1. NAMAA combined MARBERTv2, GPT-5.6, and Gemini-2.5-Flash through 2-of-3 character-level agreement. It obtained the highest overlap F1, 54.48, but ranked second with 46.41 Overall because its class macro-F1 was lower at 38.34.

Two challenges affected all submitted systems. First, the categories were highly imbalanced: sociolinguistics represented approximately 58\% of the training spans, whereas graphetics appeared only twice and was absent from development. Systems therefore used focal loss, class weighting, label smoothing, category guidance, and calibration, but rare classes remained difficult. Second, precise localization was a major bottleneck. Generative systems struggled with offsets, while token classifiers were constrained by subword boundaries; overly broad post-processing improved overlap but harmed exact match. Thakaa found that category accuracy reached 75.35\% once a span was matched, indicating that discovering and localizing errors was often harder than classifying them. Overall, the strongest systems assigned semantic proposal to large generative models and retained span verification and boundary control with dialect-aware encoders.

Table~\ref{tab:alexandriax_subtask3} presents the official results for Subtask~3 submissions alongside the baseline results, while Tables~\ref{tab:subtask3_span_results} and \ref{tab:subtask3_class_results} show the dialect breakdown for span localization and error classification, respectively.

%% file: tables/participants.tex
\begin{table*}[ht]
\centering
\small
\renewcommand{\arraystretch}{1.4} 
\resizebox{0.97\textwidth}{!}{
\begin{tabularx}{\textwidth}{@{} l X c c c c @{}}
\toprule
\multirow{2}{*}{\textbf{Team Name}} & 
\multirow{2}{*}{\textbf{Affiliation}} & 
\multicolumn{2}{c}{\textbf{Subtask 1}} & 
\multirow{2}{*}{\textbf{Subtask 2}} & 
\multirow{2}{*}{\textbf{Subtask 3}} \\
\cmidrule(lr){3-4}
& & \textbf{C} & \textbf{U} & & \\
\midrule

\rowcolor{gray!15}
ALEXIS \cite{ibrahim-etal-2026-alexis} & Alexandria University & & & \checkmark & \\

Alkhder \cite{alkhder-etal-2026-context} & Sakarya University & \checkmark & \checkmark &  &  \\

\rowcolor{gray!15}

ArabicMTDiagnostics \cite{Sarah-2026-ArabicMTDiagnostics} & Lebanese University & & & & \checkmark \\

Axiom \cite{hossain-2026-axiom} & CUET & & & & \checkmark \\

\rowcolor{gray!15}

CUNI \cite{jon-bojar-2026-cuni} & CUNI & & \checkmark & & \\

FCDS \cite{aly-etal-2026-fcds} & Alexandria University & \checkmark & & & \\

\rowcolor{gray!15}
L3IA & USMBA & & & \checkmark & \\

LahjaMT \cite{abdallah-elbeltagy-2026-lahjamt} & Newgiza University, Nawy AI lab & \checkmark & \checkmark & & \\

\rowcolor{gray!15}
NAMAA Community \cite{eldin-etal-2026-namaa} & NAMAA & \checkmark & \checkmark & \checkmark & \checkmark \\

REGLAT \cite{labib-etal-2026-reglat} & Elsewedy University of Technology & & & & \checkmark \\

\rowcolor{gray!15}
Rosetta \cite{esmaeil-etal-2026-rosetta} & Tanta University & \checkmark & \checkmark & & \\

Thakaa \cite{alamr-etal-2026-thakaa} & Thakaa & \checkmark & \checkmark & \checkmark & \checkmark \\

\rowcolor{gray!15}
TTLab \cite{abusaleh-etal-2026-ttlab} & Goethe University Frankfurt & & & & \checkmark \\

\bottomrule
\end{tabularx}}
\caption{AlexandriaX~2026 test phase participants by subtask and track. \checkmark\ denotes submitted predictions; uncited teams did not submit a system paper. For Subtask~1, \textbf{C} and \textbf{U} indicate constrained and unconstrained tracks.}
\label{tab:team-participation}
\end{table*}

%% file: tables/results.tex
\begin{table}[ht]
\centering
\small
\resizebox{0.48\textwidth}{!}{
\begin{tabular}{clcc}
\toprule
\textbf{Rank} & \textbf{Team} & \textbf{Average spBLEU} & \textbf{Average chrF++} \\
\midrule
1 & Thakaa             & 33.49 & 48.13 \\
2 & CUNI             & 32.50 & 47.05 \\
3 & LahjaMT           & 28.54 & 44.02 \\
4 & NAMAA Community   & 27.41 & 42.58 \\
5 & Rosetta           & 25.09 & 41.02 \\
\midrule
-- & Baseline   & 21.17 & 37.45 \\
\midrule
6 & Alkhder        & 20.36 & 36.64 \\
\bottomrule
\end{tabular}}
\caption{Official results for the unconstrained track of Subtask 1. Systems are ranked by mean spBLEU across the 13 target Arabic varieties; mean chrF++ is reported as a secondary metric. Alkhder submitted the same constrained system to both tracks, hence its identical scores in Table~\ref{tab:alexandriax_subtask1_constrained}.}
\label{tab:alexandriax_subtask1_unconstrained}
\end{table}

\begin{table}[ht]
\centering
\small
\resizebox{0.48\textwidth}{!}{
\begin{tabular}{clcc}
\toprule
\textbf{Rank} & \textbf{Team} & \textbf{Average spBLEU} & \textbf{Average chrF++} \\
\midrule
1 & Thakaa             & 30.42 & 45.42 \\
2 & LahjaMT           & 28.30 & 43.81 \\
3 & FCDS              & 26.48 & 42.30 \\
4 & Rosetta           & 26.10 & 41.79 \\
5 & NAMAA Community   & 23.26 & 39.03 \\
\midrule
-- & Baseline   & 21.17 & 37.45 \\
\midrule
6 & Alkhder         & 20.36 & 36.64 \\
\bottomrule
\end{tabular}}
\caption{Official results for the constrained track of Subtask 1. Systems are ranked by mean spBLEU across the 13 target Arabic varieties; mean chrF++ is reported as a secondary metric.}
\label{tab:alexandriax_subtask1_constrained}
\end{table}

\begin{table}[ht]
\centering
\small
\resizebox{0.48\textwidth}{!}{
\begin{tabular}{clcc}
\toprule
\textbf{Rank} & \textbf{Team} & \textbf{Global spBLEU} & \textbf{Global chrF++} \\
\midrule
1 & Thakaa             & 28.40 & 42.15 \\
2 & ALEXIS            & 25.84 & 39.02 \\
3 & NAMAA Community   & 24.53 & 39.04 \\
4 & L3IA              & 14.07 & 29.75 \\
\midrule
-- & Baseline (Zero-Shot Gemma-4-31B) & 24.69 & 38.44 \\
-- & Baseline (Leave-as-is) & 12.81 & 27.31 \\
\bottomrule
\end{tabular}}
\caption{Official results for Subtask 2 (global corpus-level spBLEU and chrF++ evaluated across all 3,574 blind test instances on CodaBench).}
\label{tab:alexandriax_subtask2}
\end{table}

\begin{table*}[ht]
\centering
\small
\resizebox{0.96\textwidth}{!}{
\begin{tabular}{clcccc}
\toprule
\textbf{Rank} & \textbf{Team} &
\textbf{Overall Score} &
\textbf{Exact Match F\textsubscript{1}} &
\textbf{Overlap F\textsubscript{1}} &
\textbf{Error Class Macro-F\textsubscript{1}} \\
\midrule
1 & Thakaa                & \textbf{49.82} & \textbf{23.19} & 54.26 & \textbf{45.38} \\
2 & NAMAA Community      & 46.41 & 22.39 & \textbf{54.48} & 38.34 \\
3 & TTLab                & 40.91 & 19.40 & 46.00 & 35.82 \\
4 & REGLAT               & 40.21 & 0.41  & 42.72 & 37.70 \\
5 & Axiom                & 39.26 & 16.64 & 48.13 & 30.40 \\
6 & ArabicMTDiagnostics  & 34.84 & 12.22 & 38.18 & 31.51 \\ \midrule
-  & Baseline  & 24.29 & 2.94  & 28.39 & 20.18 \\
\bottomrule
\end{tabular}}
\caption{Official results for Subtask 3. Systems are ranked by the Overall Score, computed as the mean of Overlap F1 and Error Class Macro-F1. Exact Match F1 is reported as an additional measure of span-localization accuracy.}
\label{tab:alexandriax_subtask3}
\end{table*}

%% file: tables/subtask1_uncost_breakdown.tex
\begin{table*}[t]
\centering
\resizebox{\textwidth}{!}{%
\begin{tabular}{lrrrrrrrrrrrrr|r}
\toprule
\textbf{Team} &
\textbf{EG} &
\textbf{JO} &
\textbf{LB} &
\textbf{LY} &
\textbf{MA} &
\textbf{MR} &
\textbf{OM} &
\textbf{PS} &
\textbf{SA} &
\textbf{SD} &
\textbf{SY} &
\textbf{TN} &
\textbf{YE} &
\textbf{Avg.} \\
\midrule

Thakaa
& 37.54 & 38.80 & \textbf{35.75} & 29.45 & 27.57 & \textbf{20.18} & \textbf{38.10}
& \textbf{36.04} & \textbf{36.95} & 30.32 & \textbf{42.92} & \textbf{33.16} & \textbf{28.57} & \textbf{33.49} \\

CUNI
& \textbf{38.29} & \textbf{40.10} & 34.69 & \textbf{30.85} & \textbf{28.11} & 18.79 & 29.31
& 35.36 & 36.05 & \textbf{32.38} & 38.68 & 32.79 & 27.12 & 32.50 \\

LahjaMT
& 32.59 & 34.93 & 31.81 & 24.70 & 24.29 & 18.00 & 28.61
& 31.69 & 31.39 & 25.53 & 37.02 & 29.62 & 20.85 & 28.54 \\

NAMAA Community
& 32.52 & 33.92 & 29.53 & 23.75 & 22.83 & 15.89 & 26.18
& 30.26 & 30.80 & 25.66 & 35.27 & 27.25 & 22.51 & 27.41 \\

Rosetta
& 30.41 & 32.26 & 28.09 & 20.14 & 21.84 & 10.83 & 24.78
& 29.56 & 29.58 & 20.49 & 33.64 & 24.50 & 20.11 & 25.09 \\
\midrule
Baseline
& 26.22 & 27.61 & 24.34 & 18.18 & 14.97 & 9.83 & 20.33
& 24.27 & 25.22 & 18.47 & 29.34 & 19.65 & 16.73 & 21.17 \\
\midrule
Alkhder
& 25.69 & 26.57 & 21.92 & 17.82 & 18.10 & 10.19 & 19.08
& 21.57 & 22.56 & 17.21 & 26.45 & 21.76 & 15.79 & 20.36 \\

\bottomrule
\end{tabular}%
}
\caption{Subtask 1 spBLEU scores by target Arabic variety for the unconstrained track. Avg. is the arithmetic mean across the 13 varieties. Alkhder's scores match Table~\ref{tab:s1_cons_spbleu_results} because the same system was submitted to both tracks.}
\label{tab:s1_uncons_spbleu_results}
\end{table*}

%% file: tables/subtask1_const_breakdown.tex
\begin{table*}[t]
\centering

\resizebox{\textwidth}{!}{%
\begin{tabular}{lrrrrrrrrrrrrr|r}
\toprule
\textbf{Team} &
\textbf{EG} &
\textbf{JO} &
\textbf{LB} &
\textbf{LY} &
\textbf{MA} &
\textbf{MR} &
\textbf{OM} &
\textbf{PS} &
\textbf{SA} &
\textbf{SD} &
\textbf{SY} &
\textbf{TN} &
\textbf{YE} &
\textbf{Avg.} \\
\midrule
Thakaa
& \textbf{33.07} & \textbf{36.17} & \textbf{32.44} & \textbf{26.72} & 24.14 & 17.67 & \textbf{35.51}
& \textbf{33.22} & \textbf{32.86} & \textbf{28.17} & \textbf{39.75} & 29.41 & \textbf{26.37} & \textbf{30.42} \\

LahjaMT
& 32.59 & 34.93 & 31.81 & 23.32 & \textbf{24.29} & \textbf{18.00} & 28.61
& 31.69 & 31.39 & 23.81 & 37.02 & \textbf{29.62} & 20.85 & 28.30 \\

FCDS
& 29.89 & 33.30 & 30.22 & 21.31 & 20.64 & 17.27 & 26.95
& 30.44 & 29.97 & 19.19 & 35.34 & 27.63 & 22.12 & 26.48 \\

Rosetta
& 30.85 & 33.39 & 29.70 & 19.82 & 21.42 & 12.13 & 26.07
& 30.55 & 30.56 & 21.59 & 34.73 & 26.12 & 22.36 & 26.10 \\

NAMAA Community
& 26.82 & 29.93 & 26.28 & 15.62 & 18.46 & 15.46 & 23.12
& 26.27 & 26.14 & 15.66 & 33.97 & 24.71 & 19.90 & 23.26 \\

\midrule
Baseline
& 26.22 & 27.61 & 24.34 & 18.18 & 14.97 & 9.83 & 20.33
& 24.27 & 25.22 & 18.47 & 29.34 & 19.65 & 16.73 & 21.17 \\
\midrule

Alkhder
& 25.69 & 26.57 & 21.92 & 17.82 & 18.10 & 10.19 & 19.08
& 21.57 & 22.56 & 17.21 & 26.45 & 21.76 & 15.79 & 20.36 \\
\bottomrule
\end{tabular}%
}
\caption{Subtask 1 spBLEU scores by target Arabic variety for the constrained track. Avg. is the arithmetic mean across the 13 varieties.}
\label{tab:s1_cons_spbleu_results}
\end{table*}

%% file: tables/subtask2_breakdown_target_dialect.tex
\begin{table*}[htbp]
\centering
\small
\resizebox{0.9\textwidth}{!}{
\begin{tabular}{lcccccc|c}
\toprule
\textbf{Team} & \textbf{Egyptian} & \textbf{Lebanese} & \textbf{Moroccan} & \textbf{Palestinian} & \textbf{Saudi} & \textbf{Tunisian} & \textbf{Avg.} \\
\midrule
Thakaa
    & \textbf{36.47}
    & \textbf{34.98}
    & 24.32
    & 20.48
    & 26.55
    & 16.44
    & \textbf{26.54} \\

ALEXIS
    & 13.53
    & 14.68
    & \textbf{33.59}
    & \textbf{29.35}
    & \textbf{37.55}
    & \textbf{27.23}
    & 25.99 \\

NAMAA Community
    & 33.18
    & 29.05
    & 18.57
    & 17.76
    & 25.67
    & 14.63
    & 23.14 \\

L3IA
    & 15.10
    & 14.58
    & 16.28
    & 11.51
    & 13.13
    & 7.19
    & 12.97 \\
\midrule
Baseline (Leave-as-is)
    & 13.96
    & 17.32
    & 8.90
    & 10.94
    & 12.34
    & 5.11
    & 11.43 \\
\bottomrule
\end{tabular}}
\caption{Subtask 2 spBLEU scores grouped by target dialect. Each score aggregates results from the five corresponding source dialects, and Avg. is the unweighted macro-average (arithmetic mean) of the six displayed target-dialect scores, which differs from the global corpus-level score in Table~\ref{tab:alexandriax_subtask2} due to dialect sample size disparity.}
\label{tab:subtask2_target_dialect}
\end{table*}

%% file: tables/subtask3_span_dialect_level.tex
\begin{table*}[t]
\centering

\resizebox{\textwidth}{!}{%
\begin{tabular}{lccccc|c}
\toprule

\textbf{Team} &
\textbf{EGY} &
\textbf{MAU} &
\textbf{MOR} &
\textbf{PAL} &
\textbf{UAE} &
\textbf{Avg.} \\

\midrule

Thakaa &
\textbf{35.19} / 62.79 &
17.48 / 37.78 &
\textbf{33.63} / \textbf{57.79} &
12.90 / 53.21 &
19.05 / \textbf{59.64} &
\textbf{23.65} / \textbf{54.24} \\

NAMAA Community &
29.82 / \textbf{64.86} &
\textbf{17.89} / 35.91 &
24.24 / 56.23 &
\textbf{19.51} / \textbf{55.45} &
\textbf{21.25} / 58.46 &
22.54 / 54.18 \\

TTLab &
23.93 / 58.63 &
17.78 / 30.32 &
16.16 / 46.48 &
18.00 / 38.46 &
20.00 / 51.64 &
19.17 / 45.11 \\

REGLAT &
2.02 / 43.03 &
0.00 / 36.45 &
0.00 / 44.02 &
0.00 / 44.69 &
0.00 / 45.27 &
0.40 / 42.69 \\

Axiom &
24.19 / 45.50 &
12.50 / \textbf{38.88} &
21.88 / 50.90 &
13.91 / 49.80 &
10.96 / 52.05 &
16.69 / 47.43 \\

ArabicMTDiagnostics &
10.69 / 32.98 &
16.16 / 33.75 &
6.61 / 34.87 &
8.93 / 42.69 &
17.61 / 44.09 &
12.00 / 37.68 \\

\midrule
Baseline &
3.25 / 24.56 &
0.00 / 29.88 &
2.20 / 37.21 &
2.02 / 16.63 &
6.06 / 33.88 &
2.71 / 28.43 \\

\bottomrule
\end{tabular}%
}

\caption{Subtask 3 span-localization performance by target dialect, reported as Exact Match F\textsubscript{1} / Overlap F\textsubscript{1}. Avg. is the arithmetic mean of the five dialect-specific scores. The best result for each metric and dialect is shown in bold.}
\label{tab:subtask3_span_results}

\end{table*}

%% file: tables/subtask3_class_dialect_level.tex
\begin{table}[t]
\centering

\resizebox{0.48\textwidth}{!}{%
\begin{tabular}{lccccc|c}
\toprule

\textbf{Team} &
\textbf{EGY} &
\textbf{MAU} &
\textbf{MOR} &
\textbf{PAL} &
\textbf{UAE} &
\textbf{Avg.} \\

\midrule

Thakaa &
\textbf{50.00} &
40.78 &
\textbf{42.48} &
\textbf{43.55} &
48.98 &
\textbf{45.16} \\

NAMAA Community &
40.35 &
29.27 &
36.36 &
29.27 &
\textbf{52.50} &
37.55 \\

TTLab &
37.61 &
26.67 &
36.36 &
30.00 &
44.62 &
35.05 \\

REGLAT &
36.36 &
\textbf{59.52} &
16.28 &
24.74 &
49.18 &
37.22 \\

Axiom &
32.26 &
23.21 &
31.25 &
19.13 &
42.47 &
29.66 \\

ArabicMTDiagnostics &
30.53 &
34.34 &
21.49 &
30.36 &
38.99 &
31.14 \\
\midrule
Baseline &
16.26 &
22.00 &
17.58 &
16.16 &
27.27 &
19.85 \\

\bottomrule
\end{tabular}%
}

\caption{Subtask 3 error-classification performance by target dialect, measured using Error Class Macro-F\textsubscript{1}. Avg. is the unweighted mean of the five dialect-specific scores. The best result for each dialect is shown in bold.}
\label{tab:subtask3_class_results}

\end{table}

%% file: sections/discussion.tex
\section{Discussion} \label{sec:discussion}

Across the three subtasks, several common lessons emerged. First, the strongest systems generally benefited more from targeted adaptation and careful inference strategies than from simply increasing model size or adding more data. In Subtask 1, dialect-specific adaptation, effective use of dialogue context, and candidate selection were consistently useful, while broad external-data augmentation produced mixed results. Context was most helpful when systems accounted for the mismatch between clean training histories and model-generated histories at inference time.

Subtask 2 similarly showed that retrieval, synthetic or mined supervision, and prompting can compensate for limited parallel data, but the quality and provenance of the evidence matter. Retrieved target-dialect examples were useful for guiding vocabulary, morphology, and style, whereas noisy or synthetic pairs were less reliable when treated as equivalent to human references.

Subtask 3 highlighted error localization as a major remaining challenge. Generative systems could reason well about error types but struggled with precise offsets, while token-based systems were better anchored to the input but remained sensitive to span boundaries. The strongest approaches therefore combined generative semantic reasoning with encoder-based localization. More broadly, performance varied substantially across dialects, reinforcing the need for dialect-specific modeling rather than treating Dialectal Arabic as a uniform setting.

For future editions, we see several priorities. Evaluation should move beyond single-reference lexical metrics toward multiple references or human/dialect-sensitive evaluation, since valid lexical, orthographic, and code-switching variation can be penalized by current metrics. Future tasks should also expand low-resource and subdialect coverage, and enlarge and balance the Subtask 3 annotations.

%% file: sections/conclusion.tex
\section{Conclusion} \label{sec:conclusion}

The AlexandriaX~2026 Shared Task provides a comprehensive benchmark for dialectal Arabic machine translation through three complementary subtasks. Subtask~1 evaluates context-aware translation from English into 13 Arabic varieties, using dialogue history and metadata such as domain, speaker and addressee gender, and persona information. Subtask~2 evaluates direct translation among six Arabic dialects, while Subtask~3 assesses the span-level detection and linguistic classification of machine translation errors across five English-to-dialect directions. Together, these subtasks address translation quality, cross-dialect generalization, contextual appropriateness, and interpretable error analysis.

The shared task received strong community participation from diverse international teams across the different subtasks, with competitive systems that largely outperformed the provided baselines using innovative methods including parameter-efficient and full fine-tuning, LLM prompting, ensemble and routing strategies, data augmentation, and context modeling. Nonetheless, substantial headroom remains and results across the various subtasks underscore that dialectal Arabic MT and fine-grained error localization remain open challenges. The AlexandriaX 2026 Shared Task provides a foundation for systematic evaluation and continued progress in dialectal Arabic machine translation, supporting the development of more robust and inclusive language technologies for Arabic-speaking communities across diverse dialects and contexts.

%% file: sections/limitations.tex
\section*{Limitations}

We acknowledge that the AlexandriaX 2026 Shared Task has several limitations:

\begin{itemize}

    \item \textbf{Coverage of Arabic dialects and subdialects.}
    While the task covers a broad set of Arabic varieties, it primarily models them at the country or regional level, omitting intra-dialectal and subdialectal variation across cities, age groups, and speech communities. As a result, strong country-level performance may not generalize across all speakers of a given variety.

    \item \textbf{Uneven dialect and domain coverage across subtasks.}
    Dialect and domain distributions vary significantly across subtasks: Subtask~1 covers 13 varieties across multiple domains, Subtask~2 targets fewer varieties in the banking and finance domain, and Subtask~3 evaluates five English-to-dialect directions. Performance here may not reflect broader dialectal MT capabilities across other genres or settings.

    \item \textbf{Limitations of reference-based evaluation.}
    Subtasks~1 and~2 rely on automatic reference-based metrics (spBLEU and chrF++), which struggle to capture semantic adequacy, pragmatic nuance, and valid dialectal variation. This is especially problematic given the high lexical and orthographic variance inherent to Dialectal Arabic. Furthermore, these metrics exclude key natural phenomena such as code-switching and dialectal language mixing.

    \item \textbf{Limited coverage of fine-grained linguistic phenomena.}
    Although Subtask~3 offers span-level error detection and LQM-based classification, it is constrained by a small test set of only five varieties and six linguistic categories.

\end{itemize}

%% file: sections/acknowledgments.tex
\section*{Acknowledgments}
Muhammad Abdul-Mageed acknowledges support from Canada Research Chairs (CRC), the Natural Sciences and Engineering Research Council of Canada (NSERC; RGPIN-2026-07098), the Social Sciences and Humanities Research Council of Canada (SSHRC; 895-2020-1004), Canada Foundation for Innovation (CFI; 37771), Digital Research Alliance of Canada,\footnote{\href{https://alliancecan.ca}{https://alliancecan.ca}} and UBC ARC-Sockeye. Mustafa Jarrar acknowledges that this work was partially supported by internal research grants from the College of Humanities and Social Sciences at Hamad Bin Khalifa University.

%% file: sections/appendix.tex
\section{Related Work} \label{sec:related-work}

\paragraph{Dialectal Arabic translation resources and benchmarks.}
Early work established the value of explicitly modeling Arabic dialects rather than relying solely on MSA resources. \citet{zbib-etal-2012-machine} constructed crowdsourced Egyptian--English and Levantine--English parallel corpora and demonstrated the benefits of dialectal training data. PADIC subsequently supported translation experiments across Maghrebi and Middle Eastern dialects aligned with MSA \citep{meftouh-etal-2015-machine}, while MADAR provided parallel coverage of 25 city-level dialects and an accompanying lexicon \citep{bouamor-etal-2018-madar}. More recently, TARJAMAT evaluated instruction-tuned LLMs across ten Arabic varieties, finding persistent limitations for varieties with scarce public resources \citep{kadaoui-etal-2023-tarjamat}. \citet{jon-etal-2026-current} extended the empirical picture through bidirectional English--Arabic evaluation on 16 MADAR dialects using automatic metrics and a small-scale human evaluation. Together, these studies motivate evaluating dialectal competence explicitly rather than inferring it from performance on standard Arabic.

Shared evaluations have also addressed dialectal translation. NADI 2023 introduced dialect-to-MSA translation subtasks \citep{abdul-mageed-etal-2023-nadi}, and OSACT6 organized a dedicated dialect-to-MSA translation shared task \citep{elneima-etal-2024-osact6}. AMIYA 2026 evaluated dialectal language modeling across five country varieties, including translation between dialectal Arabic and MSA or English alongside instruction following and generation \citep{robinson-etal-2026-amiya}. AlexandriaX builds on these efforts through a coordinated evaluation of context-conditioned English-to-dialect translation, direct cross-dialect translation, and automatic linguistic error analysis.

\paragraph{Conversational context and dialect-conditioned generation.}
The importance of context extends beyond Arabic. In English--Russian subtitle translation, \citet{voita-etal-2019-good} identified deixis, ellipsis, and lexical cohesion as sources of inconsistency and showed that context-aware models improve performance on targeted evaluations. This work motivates examining translation decisions that cannot be assessed adequately from isolated sentences. For dialectal Arabic, the Alexandria dataset provides human-translated, multi-turn English--Arabic conversations spanning 13 countries and 11 domains, with speaker--addressee gender annotations and contributor city metadata \citep{el-mekki-etal-2026-alexandria}. Subtask~1 uses this resource to evaluate translation conditioned on dialogue history and metadata, including gender and persona, under constrained and unconstrained resource settings. The contribution is the shared evaluation protocol and comparison of participating systems built on Alexandria, rather than the introduction of the underlying corpus.

\paragraph{Direct cross-dialect translation.}
Direct translation among Arabic dialects has precedents in PADIC's cross-dialect experiments \citep{meftouh-etal-2015-machine} and, more recently, Lahjawi, which introduced a model for translation among 15 dialects \citep{hamed-etal-2025-lahjawi}. In the financial domain, ArBanking77 localized banking queries into MSA and Palestinian Arabic for intent detection \citep{jarrar-etal-2023-arbanking77}. AraFinNLP 2024 expanded this line of work with multi-dialect intent detection and a cross-dialect translation and intent-preservation subtask \citep{malaysha-etal-2024-arafinnlp}. Subtask~2 builds on this resource lineage through an expanded ArBanking77-based benchmark covering six dialects and 30 directed translation pairs. Its blind evaluation, without dedicated parallel training or development splits covering all directions, examines generalization across dialect pairs while requiring preservation of banking intent, terminology, and numerical information.

\paragraph{Linguistically informed translation error analysis.}
Fine-grained evaluation complements aggregate translation scores by identifying where systems fail. xCOMET combines sentence-level quality estimation with error-span detection, providing localized diagnostics alongside overall scores \citep{guerreiro-etal-2024-xcomet}. For Arabic, Ara-HOPE introduces a human post-editing evaluation framework with five error categories and a decision-tree annotation protocol for dialect-to-MSA translation \citep{alabdullah-etal-2026-ara}. LQM provides a complementary taxonomy organized around sociolinguistics, pragmatics, semantics, morphosyntax, orthography, and graphetics, supported by expert span annotations of translations involving Arabic dialects \citep{magdy-etal-2026-lqm}. Subtask~3 adopts these six linguistic categories for automatic detection and classification of erroneous spans in English-to-dialect translations across five varieties. This formulation makes both localization and linguistic classification explicit evaluation targets. Taken together, the three AlexandriaX subtasks connect context-sensitive generation and cross-dialect transfer with interpretable analysis of the resulting translation errors.

\section{Evaluation and Results}
\subsection{Evaluation} \label{app:eval}

\paragraph{Subtask 2 Dialectal Breakdown and Score Aggregation Analysis} 
Table~\ref{tab:subtask2_target_dialect} presents the translation quality broken down by target dialect variety. 
We note that the official primary leaderboard score in Table~\ref{tab:alexandriax_subtask2} (e.g., 28.40 spBLEU for Thakaa) is computed as the global corpus-level (micro-averaged) metric over all 3,574 test instances pooled together on CodaBench. In contrast, the summary column in Table~\ref{tab:subtask2_target_dialect} reports the unweighted macro-average across the six target dialect categories (e.g., 26.54 spBLEU for Thakaa). This difference stems from two factors: (1) sample size imbalance across dialect categories (e.g., Egyptian contains $N=705$ instances, whereas Tunisian contains only $N=158$), and (2) dialect-specific generation difficulty. Tunisian proved the most challenging target dialect for every system except ALEXIS (16.44 spBLEU for Thakaa). In the unweighted macro-average, Tunisian receives an equal one-sixth weight (16.7\%), which pulls down the macro-average to 26.54. Thakaa's two highest-scoring target dialects (Egyptian and Lebanese, 36.47 and 34.98) together constitute nearly 40\% of the test corpus.

\subsection{Baselines} \label{app:baselines}

For Subtask~2, we established two complementary baselines:
\begin{itemize}
    \item \textbf{Leave-as-is (Identity Copy) Baseline:} This baseline outputs the source dialect sentence unmodified without any rewriting. In cross-dialect Arabic NLP, regional dialects share substantial vocabulary, morphological structures, and MSA cognates. Consequently, an unmodified input can achieve deceptively high surface overlap (e.g., exceeding 21 spBLEU between closely related varieties like Egyptian and Lebanese), creating an ``illusion of translation.'' Because Subtask~2 provided no parallel training data across all 30 evaluated dialect pairs, the \textit{Leave-as-is} baseline is essential to establish the empirical floor of passive lexical similarity (achieving a global corpus average of 12.81 spBLEU and 27.31 chrF++) and to diagnose whether submitted models execute genuine dialectal transfer or rely on copy-through shortcuts.
    \item \textbf{Zero-Shot Model Baseline:} To provide a standard model-based reference representing an off-the-shelf translation approach, zero-shot prompting with Gemma-4-31B-IT achieves 24.69 spBLEU and 38.44 chrF++ (and similarly Gemini-2.5-Flash at 24.53 spBLEU / 39.04 chrF++ as evaluated by NAMAA Community), contextualizing the explicit performance gains delivered by participants' dedicated retrieval, adaptation, and fine-tuning pipelines.
\end{itemize}